\documentclass[11pt,a4paper]{article}
\usepackage[hyperref]{acl2020}
\usepackage{times}
\usepackage{latexsym}

\usepackage{microtype}
\usepackage{booktabs}
\usepackage{amsfonts}
\usepackage{amsmath}
\usepackage{multirow}
\usepackage{url}
\PassOptionsToPackage{linktocpage}{hyperref}

\aclfinalcopy 
\def\aclpaperid{2333} 

\newcommand{\dataset}{CAIL2019-SCM}

\title{\dataset: A Dataset of Similar Case Matching in Legal Domain}

\author{
Chaojun Xiao$^{1\ast}$ 
Haoxi Zhong$^{1}$\thanks{$ $ $ $ $ $ indicates equal contribution.} $ $ 
Zhipeng Guo$^{1}$ 
Cunchao Tu$^{1}$ 
Zhiyuan Liu$^{1}$
Maosong Sun$^{1}$\\
\textbf{
Tianyang Zhang$^{2}$
Xianpei Han$^{3}$ 
Zhen Hu$^{4}$ 
Heng Wang$^{4}$ 
Jianfeng Xu$^{5}$}\\
\normalsize $^{1}$Department of Computer Science and Technology, Tsinghua University, China \\
\normalsize $^{2}$Beijing Powerlaw Intelligent Technology Co., Ltd., China \\
\normalsize $^{3}$Institute of Software, Chinese Academy of Sciences, China\\
\normalsize $^{4}$China Justice Big Data Institute\\
\normalsize $^{5}$Supreme People Court, China\\
}

\date{}

\begin{document}
\maketitle
\begin{abstract}
In this paper, we introduce \dataset, \textbf{C}hinese \textbf{AI} and \textbf{L}aw \textbf{2019} \textbf{S}imilar \textbf{C}ase \textbf{M}atching dataset. \dataset\ contains 8,964 triplets of cases published by the Supreme People's Court of China. \dataset\ focuses on detecting similar cases, and the participants are required to check which two cases are more similar in the triplets. There are 711 teams who participated in this year's competition, and the best team has reached a score of $71.88$. We have also implemented several baselines to help researchers better understand this task. The dataset and more details can be found from https://github.com/china-ai-law-challenge/CAIL2019/tree/master/scm.
\end{abstract}

\section{Introduction}

\textbf{S}imilar \textbf{C}ase \textbf{M}atching (SCM) plays a major role in legal system, especially in common law legal system. The most similar cases in the past determine the judgment results of cases in common law systems. As a result, legal professionals often spend much time finding and judging similar cases to prove fairness in judgment. As automatically finding similar cases can benefit to the legal system, we select SCM as one of the tasks of CAIL2019.

\textbf{C}hinese \textbf{AI} and \textbf{L}aw Challenge (CAIL) is a competition of applying artificial intelligence technology to legal tasks. The goal of the competition is to use AI to help the legal system. CAIL was first held in 2018, and the main task of CAIL2018~\cite{xiao2018cail2018,zhong2018overview} is predicting the judgment results from the fact description. The judgment results include the accusation, applicable articles, and the term of penalty. CAIL2019 contains three different tasks, including Legal Question-Answering, Legal Case Element Prediction, and Similar Case Matching. Furthermore, we will focus on SCM in this paper.

More specifically, \dataset\ contains 8,964 triplets of legal documents. Every legal documents is collected from China Judgments Online\footnote{\url{http://wenshu.court.gov.cn/}}. In order to ensure the similarity of the cases in one triplet, all selected documents are related to Private Lending. Every document in the triplet contains the fact description. \dataset\ requires researchers to decide which two cases are more similar in a triplet. By detecting similar cases in triplets, we can apply this algorithm for ranking all documents to find the most similar document in the database. There are 247 teams who have participated \dataset, and the best team has reached a score of $71.88$, which is about $20$ points higher than the baseline. The results show that the existing methods have made great progress on this task, but there is still much room for improvement.

In other words, \dataset\ can benefit the research of legal case matching. Furthermore, there are several main challenges of \dataset: (1) The difference between documents may be small, and then it is hard to decide which two documents are more similar. Moreover, the similarity is defined by legal workers. We must utilize legal knowledge into this task rather than calculate similarity on the lexical level. (2) The length of the documents is quite long. Most documents contain more than $512$ characters, and then it is hard for existing methods to capture document level information.

In the following parts, we will give more details about \dataset, including related works about SCM, the task definition, the construction of the dataset, and several experiments on the dataset.

\section{Related Work}
\subsection{Semantic Text Matching}
SCM aims to measure the similarity between legal case documents. Essentially, it is an application of semantic text matching, which is central for many tasks in natural language processing, such as question answering, information retrieval, and natural language inference. Take information retrieval as an example, given a query and a database, a semantic matching model is required to judge the semantic similarity between the query and documents in the database. Moreover, the tasks related to semantic matching have attracted the attention of many researchers in recent decades.

Intuitively traditional approaches calculate word-to-word similarity with vector space model, e.g. term frequency-inverse document frequency~\cite{wu2008interpreting}, bag-of-words~\cite{bilotti2007structured}. However, due to the variety of words in different texts, these approaches achieve limited success in the task.

Recently, with the development of deep learning in natural language processing, researchers attempt to apply neural models to encode text into distributed representation. The Siamese structure~\cite{bromley1994signature} for metric learning achieve great success and is widely applied~\cite{amiri2016learning,liu2018matching,mueller2016siamese,neculoiu2016learning,wan2016deep,he2015multi}. Besides, there are many researchers put emphasis on integrating syntactic structure into semantic matching~\cite{liu2018matching,chen2017enhanced} and multi-level text matching with attention-aware representation~\cite{duan2018attention,tan2018multiway,yin2016abcnn}.

Nevertheless, most previous studies are designed for identifying the relationship between two sentences with limited length.  

\subsection{Legal Intelligence}
Researchers widely concern tasks for legal intelligence. Applying NLP techniques to solve a legal problem becomes more and more popular in recent years. Previous works~\cite{kort1957predicting,keown1980mathematical,lauderdale2012supreme} focus on analyzing existing cases with mathematical tools. With the development of deep learning, more researchers pay much efforts on predicting the judgment result of legal cases~\cite{luo2017learning,hufew,zhong2018legal,chalkidis2019neural,jiang2018interpretable,yanglegal}. Furthermore, there are many works on generating court views to interpret charge results~\cite{ye2018interpretable}, information extraction from legal text~\cite{vacek2017sequence,vacek2019litigation}, legal event detection~\cite{yan2017event}, identifying applicable law articles~\cite{liu2015predicting,liu2006exploring} and legal question answering~\cite{kim2015coliee,fawei2018methodology}. 

Meanwhile, retrieving related legal documents with a query has been studied for decades and is a critical issue in applications of legal intelligence. 
\citet{raghav2016analyzing} emphasize exploiting paragraph-level and citation information. 
\citet{kano2017overview} and \citet{zhong2018overview} held a legal information extraction and entailment competition to promote progress in legal case retrieval.

\section{Overview of Dataset}

\subsection{Task Definition}

We first define the task of \dataset\ here. The input of \dataset\ is a triplet $(A,B,C)$, where $A,B,C$ are fact descriptions of three cases. Here we define a function $sim$ which is used for measuring the similarity between two cases. Then the task of \dataset\ is to predict whether $sim(A,B)>sim(A,C)$ or $sim(A,C)>sim(A,B)$.

\subsection{Dataset Construction and Details}

To ensure the quality of the dataset, we have several steps of constructing the dataset. First, we select many documents within the range of Private Lending. However, although all cases are related to Private Lending, they are still various so that many cases are not similar at all. If the cases in the triplets are not similar, it does not make sense to compare their similarities. To produce qualified triplets, we first annotated some crucial elements in Private Lending for each document. The elements include:

\begin{itemize}
    \item The properties of lender and borrower, whether they are a natural person, a legal person, or some other organization.
    \item The type of guarantee, including no guarantee, guarantee, mortgage, pledge, and others.
    \item The usage of the loan, including personal life, family life, enterprise production and operation, crime, and others.
    \item The lending intention, including regular lending, transfer loan, and others.
    \item Conventional interest rate method, including no interest, simple interest, compound interest, unclear agreement, and others.
    \item Interest during the agreed period, including $[0\%,24\%]$,  $(24\%,36\%]$, $(36\%,\infty)$, and others.
    \item Borrowing delivery form, including no lending, cash, bank transfer, online electronic remittance, bill, online loan platform, authorization to control a specific fund account, unknown or fuzzy, and others.
    \item Repayment form, including unpaid, partial repayment, cash, bank transfer, online electronic remittance, bill, unknown or fuzzy, and others.
    \item Loan agreement, including loan contract, or borrowing, ``WeChat, SMS, phone or other chat records'', receipt, irrigation, repayment commitment, guarantee, unknown or fuzzy and others.
\end{itemize}

After annotating these elements, we can assume that cases with similar elements are quite similar. So when we construct the triplets, we calculate the tf-idf similarity and elemental similarity between cases and select those similar cases to construct our dataset. We have constructed 8,964 triples in total by these methods, and the statistics can be found from Table~\ref{tab:statistics}. Then, legal professionals will annotate every triplet to see whether $sim(A,B)>sim(A,C)$ or $sim(A,B)<sim(A,C)$. Furthermore, to ensure the quality of annotation, every document and triplet is annotated by at least three legal professionals to reach an agreement.

\begin{table}[h]
    \centering
    \begin{tabular}{c|c} \toprule
         Type &  Count \\ \midrule
         Small Train & 500 \\
         Small Test & 326\\\midrule
         Large Train & 5,102 \\
         Large Valid & 1,500\\
         Large Test & 1,536 \\\midrule
         Total & 8,964\\\bottomrule
    \end{tabular}
    \caption{The number of triplets in different stages of \dataset.}
    \label{tab:statistics}
\end{table}

\section{Experiments}
To access the challenge of the similar cases matching task, we evaluate several baselines on our dataset. The experiment results show that even the state-of-the-art systems perform poorly in evaluating the similarity between different cases. 

\textbf{Baselines.} All the baseline models are trained on \textit{Large Train} and tested on \textit{Large Valid} and \textit{Large Test}. We adapt the Siamese framework~\cite{bromley1994signature} to our scenario with different encoder, e.g. CNN~\cite{kim2014convolutional}, LSTM~\cite{hochreiter1997long}, Bert~\cite{devlin2019bert}, used for encoding the legal documents. We will elaborate on the details of the framework in the following part.

Given the triplet of fact description, ($D_A$, $D_B$, $D_C$), we first encode them into distributed vectors with the same encoder and then compute the similarity scores between the query case $D_A$ and the candidate cases $D_B$, $D_C$ with a linear layer. Assume that each document $D$ consisting of $n$ words, i.e. $D = \{w_1, w_2, ..., w_n\}$. 

For CNN/LSTM encoder, we first employ THULAC~\cite{sun2016thulac} for word segmentation and then  transform each word into distributed representation $X = \{x_1, x_2, ..., x_n\}$ with Glove \cite{pennington2014glove}, where $x_i \in \mathbb{R}^d, i = 1, 2, ..., n$ and $d$ is the dimension of word embeddings. Next, the encoder layer and max pooling layer transform the embedding sequence $X$ into features $h \in \mathbb{R}^{d_h}$, where $d_h$ is the dimension of hidden vector. While for Bert encoder, we feed the document in character-level into the $bert\_base\_chinese$ model to get the features $h$.
\begin{equation}
\begin{aligned}
    h_A &= Encoder(D_A) \\
    h_B &= Encoder(D_B) \\
    h_C &= Encoder(D_C)
\end{aligned}
\end{equation}

Afterward, we calculate the similarity with a linear layer with softmax activation. $W \in \mathbb{R}^{d_h \times d_h}$ is a weight matrix to be learned.
\begin{equation}
\begin{aligned}
    s_{Aj} &= \text{softmax}(exp(h_A W h_j)) \\
    j &= B,C
\end{aligned}
\end{equation}

For the learning objective, we apply the binary cross-entropy loss function with ground-truth label $p$:
\begin{equation}
    L(\theta) = \mathbb{E}[p \text{ln}(s_{AB}) + (1 - p) \text{ln}(s_{AC})]
\end{equation}

\begin{table}[]
\centering
\begin{tabular}{l|l|ll} \toprule
                           & Method     & Valid   & Test  \\ \midrule
\multirow{3}{*}{baselines} & CNN        & 62.27 & 69.53 \\
                           & LSTM       & 62.00 & 68.00 \\
                           & BERT       & 61.93 & 67.32 \\ \midrule
\multirow{3}{*}{Teams}     & AlphaCourt & \textbf{70.07} & \textbf{72.66} \\
                           & backward   & 67.73 & 71.81 \\
                           & 11.2 yuan      & 66.73 & 72.07 \\ \bottomrule
\end{tabular}
\caption{Results of baselines and scores of top 3 participants on valid and test datasets.}
\label{tab:results}
\end{table}

\textbf{Model Performance.} We use the accuracy metric in our experiments. Table~\ref{tab:results} shows the results of baselines and top 3 participant teams on \textit{Large Valid} and \textit{Large Test} dataset, from which we get the following conclusion: 1) The participants achieve promising progress compared to baseline models. 2) Both the baselines systems and participant teams perform poorly on the dataset, due to the lack of utilization of prior legal knowledge. It's still challenging to utilize legal knowledge and simulate legal reasoning for the dataset.

\section{Conclusion}
In this paper, we propose a new dataset, \dataset, which focuses on the task of similar case matching in the legal domain. Compared with existing datasets, \dataset\ can benefit the case matching in the legal domain to help the legal partitioners work better. Experimental results also show that there is still plenty of room for improvement.

\bibliography{acl2020}

\begin{thebibliography}{39}
\expandafter\ifx\csname natexlab\endcsname\relax\def\natexlab#1{#1}\fi

\bibitem[{Amiri et~al.(2016)Amiri, Resnik, Boyd-Graber, and
  Daum{\'e}~III}]{amiri2016learning}
Hadi Amiri, Philip Resnik, Jordan Boyd-Graber, and Hal Daum{\'e}~III. 2016.
\newblock Learning text pair similarity with context-sensitive autoencoders.
\newblock In \emph{Proceedings of the 54th Annual Meeting of the Association
  for Computational Linguistics (Volume 1: Long Papers)}, pages 1882--1892.

\bibitem[{Bilotti et~al.(2007)Bilotti, Ogilvie, Callan, and
  Nyberg}]{bilotti2007structured}
Matthew~W Bilotti, Paul Ogilvie, Jamie Callan, and Eric Nyberg. 2007.
\newblock Structured retrieval for question answering.
\newblock In \emph{Proceedings of the 30th annual international ACM SIGIR
  conference on Research and development in information retrieval}, pages
  351--358. ACM.

\bibitem[{Bromley et~al.(1994)Bromley, Guyon, LeCun, S{\"a}ckinger, and
  Shah}]{bromley1994signature}
Jane Bromley, Isabelle Guyon, Yann LeCun, Eduard S{\"a}ckinger, and Roopak
  Shah. 1994.
\newblock Signature verification using a" siamese" time delay neural network.
\newblock In \emph{Advances in neural information processing systems}, pages
  737--744.

\bibitem[{Chalkidis et~al.(2019)Chalkidis, Androutsopoulos, and
  Aletras}]{chalkidis2019neural}
Ilias Chalkidis, Ion Androutsopoulos, and Nikolaos Aletras. 2019.
\newblock Neural legal judgment prediction in english.
\newblock \emph{In Proceddings of ACL}.

\bibitem[{Chen et~al.(2017)Chen, Zhu, Ling, Wei, Jiang, and
  Inkpen}]{chen2017enhanced}
Qian Chen, Xiaodan Zhu, Zhen-Hua Ling, Si~Wei, Hui Jiang, and Diana Inkpen.
  2017.
\newblock Enhanced lstm for natural language inference.
\newblock In \emph{Proceedings of the 55th Annual Meeting of the Association
  for Computational Linguistics (Volume 1: Long Papers)}, pages 1657--1668.

\bibitem[{Devlin et~al.(2019)Devlin, Chang, Lee, and
  Toutanova}]{devlin2019bert}
Jacob Devlin, Ming-Wei Chang, Kenton Lee, and Kristina Toutanova. 2019.
\newblock Bert: Pre-training of deep bidirectional transformers for language
  understanding.
\newblock In \emph{Proceedings of the 2019 Conference of the North American
  Chapter of the Association for Computational Linguistics: Human Language
  Technologies, Volume 1 (Long and Short Papers)}, pages 4171--4186.

\bibitem[{Duan et~al.(2018)Duan, Cui, Chen, Wei, Zhu, and
  Zhao}]{duan2018attention}
Chaoqun Duan, Lei Cui, Xinchi Chen, Furu Wei, Conghui Zhu, and Tiejun Zhao.
  2018.
\newblock Attention-fused deep matching network for natural language inference.
\newblock In \emph{IJCAI}, pages 4033--4040.

\bibitem[{Fawei et~al.(2018)Fawei, Pan, Kollingbaum, and
  Wyner}]{fawei2018methodology}
Biralatei Fawei, Jeff~Z Pan, Martin Kollingbaum, and Adam~Z Wyner. 2018.
\newblock A methodology for a criminal law and procedure ontology for legal
  question answering.
\newblock In \emph{In Proceddings of JIST}. Springer Verlag.

\bibitem[{He et~al.(2015)He, Gimpel, and Lin}]{he2015multi}
Hua He, Kevin Gimpel, and Jimmy Lin. 2015.
\newblock Multi-perspective sentence similarity modeling with convolutional
  neural networks.
\newblock In \emph{Proceedings of the 2015 Conference on Empirical Methods in
  Natural Language Processing}, pages 1576--1586.

\bibitem[{Hochreiter and Schmidhuber(1997)}]{hochreiter1997long}
Sepp Hochreiter and J{\"u}rgen Schmidhuber. 1997.
\newblock Long short-term memory.
\newblock \emph{Neural computation}, 9(8):1735--1780.

\bibitem[{Hu et~al.(2018)Hu, Li, Liu, and Sun}]{hufew}
Zikun Hu, Xiang Li, Cunchao Tu~Zhiyuan Liu, and Maosong Sun. 2018.
\newblock Few-shot charge prediction with discriminative legal attributes.

\bibitem[{Jiang et~al.(2018)Jiang, Ye, Luo, Chao, and
  Ma}]{jiang2018interpretable}
Xin Jiang, Hai Ye, Zhunchen Luo, WenHan Chao, and Wenjia Ma. 2018.
\newblock Interpretable rationale augmented charge prediction system.
\newblock In \emph{In Proceedings of COLING}.

\bibitem[{Kano et~al.(2017)Kano, Kim, Goebel, and Satoh}]{kano2017overview}
Yoshinobu Kano, Mi-Young Kim, Randy Goebel, and Ken Satoh. 2017.
\newblock Overview of coliee 2017.
\newblock In \emph{COLIEE@ ICAIL}, pages 1--8.

\bibitem[{Keown(1980)}]{keown1980mathematical}
R~Keown. 1980.
\newblock Mathematical models for legal prediction.
\newblock \emph{Computer/lj}, 2:829.

\bibitem[{Kim et~al.(2015)Kim, Goebel, and Ken}]{kim2015coliee}
Mi-Young Kim, Randy Goebel, and S~Ken. 2015.
\newblock Coliee-2015: evaluation of legal question answering.
\newblock In \emph{In Proceddings of JURISIN}.

\bibitem[{Kim(2014)}]{kim2014convolutional}
Yoon Kim. 2014.
\newblock Convolutional neural networks for sentence classification.
\newblock In \emph{Proceedings of the 2014 Conference on Empirical Methods in
  Natural Language Processing (EMNLP)}, pages 1746--1751.

\bibitem[{Kort(1957)}]{kort1957predicting}
Fred Kort. 1957.
\newblock Predicting supreme court decisions mathematically: A quantitative
  analysis of the" right to counsel" cases.
\newblock \emph{The American Political Science Review}, 51(1):1--12.

\bibitem[{Lauderdale and Clark(2012)}]{lauderdale2012supreme}
Benjamin~E Lauderdale and Tom~S Clark. 2012.
\newblock The supreme court's many median justices.
\newblock \emph{American Political Science Review}, 106(4):847--866.

\bibitem[{Liu et~al.(2018)Liu, Zhang, Han, Niu, Lai, and Xu}]{liu2018matching}
Bang Liu, Ting Zhang, Fred~X Han, Di~Niu, Kunfeng Lai, and Yu~Xu. 2018.
\newblock Matching natural language sentences with hierarchical sentence
  factorization.
\newblock In \emph{Proceedings of the 2018 World Wide Web Conference}, pages
  1237--1246. International World Wide Web Conferences Steering Committee.

\bibitem[{Liu and Hsieh(2006)}]{liu2006exploring}
Chao-Lin Liu and Chwen-Dar Hsieh. 2006.
\newblock Exploring phrase-based classification of judicial documents for
  criminal charges in chinese.
\newblock In \emph{Proceedings of the 16th international conference on
  Foundations of Intelligent Systems}. Springer-Verlag.

\bibitem[{Liu et~al.(2015)Liu, Chen, and Ho}]{liu2015predicting}
Yi-Hung Liu, Yen-Liang Chen, and Wu-Liang Ho. 2015.
\newblock Predicting associated statutes for legal problems.
\newblock \emph{Information Processing \& Management}, 51(1):194--211.

\bibitem[{Luo et~al.(2017)Luo, Feng, Xu, Zhang, and Zhao}]{luo2017learning}
Bingfeng Luo, Yansong Feng, Jianbo Xu, Xiang Zhang, and Dongyan Zhao. 2017.
\newblock Learning to predict charges for criminal cases with legal basis.
\newblock In \emph{In Proceedings of EMNLP}.

\bibitem[{Mueller and Thyagarajan(2016)}]{mueller2016siamese}
Jonas Mueller and Aditya Thyagarajan. 2016.
\newblock Siamese recurrent architectures for learning sentence similarity.
\newblock In \emph{Thirtieth AAAI Conference on Artificial Intelligence}.

\bibitem[{Neculoiu et~al.(2016)Neculoiu, Versteegh, and
  Rotaru}]{neculoiu2016learning}
Paul Neculoiu, Maarten Versteegh, and Mihai Rotaru. 2016.
\newblock Learning text similarity with siamese recurrent networks.
\newblock In \emph{Proceedings of the 1st Workshop on Representation Learning
  for NLP}, pages 148--157.

\bibitem[{Pennington et~al.(2014)Pennington, Socher, and
  Manning}]{pennington2014glove}
Jeffrey Pennington, Richard Socher, and Christopher Manning. 2014.
\newblock Glove: Global vectors for word representation.
\newblock In \emph{Proceedings of the 2014 conference on empirical methods in
  natural language processing (EMNLP)}, pages 1532--1543.

\bibitem[{Raghav et~al.(2016)Raghav, Reddy, and Reddy}]{raghav2016analyzing}
K~Raghav, P~Krishna Reddy, and V~Balakista Reddy. 2016.
\newblock Analyzing the extraction of relevant legal judgments using
  paragraph-level and citation information.
\newblock \emph{AI4JCArtificial Intelligence for Justice}, page~30.

\bibitem[{Sun et~al.(2016)Sun, Chen, Zhang, Guo, and Liu}]{sun2016thulac}
Maosong Sun, Xinxiong Chen, Kaixu Zhang, Zhipeng Guo, and Zhiyuan Liu. 2016.
\newblock Thulac: An efficient lexical analyzer for chinese.
\newblock Technical report, Technical Report. Technical Report.

\bibitem[{Tan et~al.(2018)Tan, Wei, Wang, Lv, and Zhou}]{tan2018multiway}
Chuanqi Tan, Furu Wei, Wenhui Wang, Weifeng Lv, and Ming Zhou. 2018.
\newblock Multiway attention networks for modeling sentence pairs.
\newblock In \emph{IJCAI}, pages 4411--4417.

\bibitem[{Vacek et~al.(2019)Vacek, Teo, Song, Cowling, Schilder, Nugent, and
  Wharf}]{vacek2019litigation}
Thomas Vacek, Ronald Teo, Dezhao Song, Conner Cowling, Frank Schilder, Timothy
  Nugent, and Canary Wharf. 2019.
\newblock Litigation analytics: Case outcomes extracted from us federal court
  dockets.
\newblock \emph{In Proceddings of NAACL-HLT}.

\bibitem[{Vacek and Schilder(2017)}]{vacek2017sequence}
Tom Vacek and Frank Schilder. 2017.
\newblock A sequence approach to case outcome detection.
\newblock In \emph{In Proceedings of ICAIL}. ACM.

\bibitem[{Wan et~al.(2016)Wan, Lan, Guo, Xu, Pang, and Cheng}]{wan2016deep}
Shengxian Wan, Yanyan Lan, Jiafeng Guo, Jun Xu, Liang Pang, and Xueqi Cheng.
  2016.
\newblock A deep architecture for semantic matching with multiple positional
  sentence representations.
\newblock In \emph{Thirtieth AAAI Conference on Artificial Intelligence}.

\bibitem[{Wu et~al.(2008)Wu, Luk, Wong, and Kwok}]{wu2008interpreting}
Ho~Chung Wu, Robert Wing~Pong Luk, Kam~Fai Wong, and Kui~Lam Kwok. 2008.
\newblock Interpreting tf-idf term weights as making relevance decisions.
\newblock \emph{ACM Transactions on Information Systems (TOIS)}, 26(3):13.

\bibitem[{Xiao et~al.(2018)Xiao, Zhong, Guo, Tu, Liu, Sun, Feng, Han, Hu, Wang
  et~al.}]{xiao2018cail2018}
Chaojun Xiao, Haoxi Zhong, Zhipeng Guo, Cunchao Tu, Zhiyuan Liu, Maosong Sun,
  Yansong Feng, Xianpei Han, Zhen Hu, Heng Wang, et~al. 2018.
\newblock Cail2018: A large-scale legal dataset for judgment prediction.
\newblock \emph{arXiv preprint arXiv:1807.02478}.

\bibitem[{Yan et~al.(2017)Yan, Zheng, Lu, and Song}]{yan2017event}
Yukun Yan, Daqi Zheng, Zhengdong Lu, and Sen Song. 2017.
\newblock Event identification as a decision process with non-linear
  representation of text.
\newblock \emph{arXiv preprint arXiv:1710.00969}.

\bibitem[{Yang et~al.(2019)Yang, Jia, Zhou, and Luo}]{yanglegal}
Wenmian Yang, Weijia Jia, Xiaojie Zhou, and Yutao Luo. 2019.
\newblock Legal judgment prediction via multi-perspective bi-feedback network.

\bibitem[{Ye et~al.(2018)Ye, Jiang, Luo, and Chao}]{ye2018interpretable}
Hai Ye, Xin Jiang, Zhunchen Luo, and Wenhan Chao. 2018.
\newblock Interpretable charge predictions for criminal cases: Learning to
  generate court views from fact descriptions.
\newblock In \emph{In Proceedings ofNAACL}.

\bibitem[{Yin et~al.(2016)Yin, Sch{\"u}tze, Xiang, and Zhou}]{yin2016abcnn}
Wenpeng Yin, Hinrich Sch{\"u}tze, Bing Xiang, and Bowen Zhou. 2016.
\newblock Abcnn: Attention-based convolutional neural network for modeling
  sentence pairs.
\newblock \emph{Transactions of the Association for Computational Linguistics},
  4:259--272.

\bibitem[{Zhong et~al.(2018{\natexlab{a}})Zhong, Guo, Tu, Xiao, Liu, and
  Sun}]{zhong2018legal}
Haoxi Zhong, Zhipeng Guo, Cunchao Tu, Chaojun Xiao, Zhiyuan Liu, and Maosong
  Sun. 2018{\natexlab{a}}.
\newblock Legal judgment prediction via topological learning.
\newblock In \emph{In Proceedings of the EMNLP}.

\bibitem[{Zhong et~al.(2018{\natexlab{b}})Zhong, Xiao, Guo, Tu, Liu, Sun, Feng,
  Han, Hu, Wang et~al.}]{zhong2018overview}
Haoxi Zhong, Chaojun Xiao, Zhipeng Guo, Cunchao Tu, Zhiyuan Liu, Maosong Sun,
  Yansong Feng, Xianpei Han, Zhen Hu, Heng Wang, et~al. 2018{\natexlab{b}}.
\newblock Overview of cail2018: Legal judgment prediction competition.
\newblock \emph{arXiv preprint arXiv:1810.05851}.

\end{thebibliography}
\bibliographystyle{acl_natbib}

\end{document}